\documentclass[table,xcdraw]{article} 
\usepackage{iclr2020_conference,times}

\usepackage{amsmath,amsfonts,bm}

\def\eqref#1{equation~\ref{#1}}

\def\1{\bm{1}}

\DeclareMathAlphabet{\mathsfit}{\encodingdefault}{\sfdefault}{m}{sl}
\SetMathAlphabet{\mathsfit}{bold}{\encodingdefault}{\sfdefault}{bx}{n}

\usepackage{hyperref}
\usepackage{url}
\usepackage{epsfig}
\usepackage{caption}
\usepackage{graphicx}
\usepackage{graphbox}
\usepackage{amsmath}
\usepackage{todonotes}
\usepackage{animate}
\usepackage{multirow}
\usepackage{subcaption}
\usepackage{floatrow}
\usepackage{float}
\usepackage{algpseudocode}
\floatstyle{plaintop}
\restylefloat{table}
\usepackage{verbatim}
\usepackage[symbol]{footmisc}

\title{Interpreting video features: A comparison of 3D Convolutional networks and Convolutional LSTM networks}

\iclrfinalcopy

\author{Joonatan M\"antt\"ari\footnotemark[1], Sofia Broom\'e\footnote[1]{Equal contribution}, John Folkesson \& Hedvig Kjellstr\"om \\
Division of Robotics, Perception and Learning\\
Department of Electrical Engineering and Computer Science\\
KTH Royal Institute of Technology\\
Stockholm, Sweden \\
\texttt{\{manttari,sbroome,johnf,hedvig\}@kth.se} \\
}

\begin{document}

\maketitle
\footnotetext[1]{Equal contribution}
\begin{abstract}
    A number of techniques for interpretability have been presented for deep learning in computer vision, typically with the goal of understanding what the networks have based their classification
    on.
    However, interpretability for deep video architectures is still in its infancy and we do not yet have a clear concept of how to decode spatiotemporal features. In this paper, we present a study comparing how 3D convolutional networks and convolutional LSTM networks learn features across temporally dependent frames. This is the first comparison of two video models that both convolve to learn spatial features but have principally different methods of modeling time. 
    Additionally, we extend the concept of meaningful perturbation introduced by \cite{MeaningFulPert} to the temporal dimension, to identify the temporal part of a sequence most meaningful to the network for a classification decision. Our findings indicate that the 3D convolutional model concentrates on shorter events in the input sequence, and places its spatial focus on fewer, contiguous areas. 
    %
\end{abstract}

\section{Introduction}
\label{section:intro}
Two standard approaches to deep learning for sequential image data are 3D convolutional neural networks (3D CNNs), e.g., ~the I3D model (\cite{I3DRef}), and recurrent neural networks (RNNs). Among the RNNs, the convolutional long short-term memory network (C-LSTM) (\cite{Shi}) is especially suited for sequences of images, since it learns both spatial and temporal dependencies simultaneously. 
  Although both methods can capture aspects of the semantics pertaining to the temporal dependencies in a video clip, there is a fundamental difference in how 3D CNNs treat time compared to C-LSTMs. In 3D CNNs, the time axis is treated just like a third spatial axis, whereas C-LSTMs only allow for information flow in the direction of increasing time, complying with the second law of thermodynamics. More concretely, C-LSTMs maintain a hidden state that is continuously updated when forward-traversing the input video sequence, and are able to model non-linear transitions in time. 3D CNNs, on the other hand, convolve (i.e., take weighted averages) over both the temporal and spatial dimensions of the sequence.
  
  The question investigated in this paper is whether this difference has consequences for how the two models compute spatiotemporal features. We present a study of how 3D CNNs and C-LSTMs respectively compute video features: what do they learn, and how do they differ from one another?

As outlined in Section 2, 
there is a large body of work on evaluating video architectures on spatial and temporal correlations, but significantly fewer investigations of what parts of the data the networks have used and what semantics relating to the temporal dependencies they have extracted from them.
Deep neural networks are known to be large computational models, whose inner workings are difficult to overview for a human. For video models, the number of parameters is typically significantly higher due to the added dimension, which makes their interpretability all the more pressing.

We will evaluate these two types of models (3D CNN and C-LSTM) on tasks where temporal order is crucial. The 20BN-Something-something-V2 dataset (\cite{Something-Something-v2}) (hereon, Something-something) will be central to our investigations; it contains time-critical classes, agnostic to object appearance, such as \textit{move something from left to right} or \textit{move something from right to left}. 
We additionally evaluate the models on the smaller KTH actions dataset (\cite{KTHDATASET}).

Our contributions are listed as follows.

\begin{itemize}
    \item We present the first comparison of 3D CNNs and C-LSTMs in terms of temporal modeling abilities. We show essential differences between their assumptions concerning temporal dependencies in the data through qualitative and quantitative experiments.
    \item We extend the concept of meaningful perturbation introduced by \cite{MeaningFulPert} to the temporal dimension, to search for the most critical part of a sequence used by the networks for classification.
\end{itemize}

\section{Related Work}

The field of interpretability in deep learning is still young but has made considerable progress for single-image networks, owing to works such as \cite{Zeiler2014}, \cite{Simonyan2014DeepMaps}, \cite{tcavglobal_icml18} and \cite{Montavon2018}.
One can distinguish between data centric and network centric methods for interpretability. \textit{Activity maximization}, first coined by \cite{Erhan2009VisualizingNetwork}, is network centric in the sense that specific units of the network are studied. By maximizing the activation of a given unit by gradient ascent with reference to the input, one can compute its optimal input. In data centric methods, the focus is instead on the input to the network in order to reveal which patterns of the data the network has discerned.

Grad-CAM (\cite{Selvaraju2017Grad-CAM:Localization}) and the meaningful perturbations explored in the work by \cite{MeaningFulPert} (Section \ref{section:approach}), which form the basis for our experiments, belong to the data centric category. Layer-wise relevance propagation (LRP, \cite{Montavon2018}) and Excitation backprop (\cite{ExcitationBackprop16}) are two other examples of data centric backpropagation techniques designed for interpretability, where Excitation backprop follows from a simpler parameter setting of LRP. In Excitation backprop, saliency maps are produced without the use of gradients. Instead, products of forward weights and activations are normalized in order to be used as conditional probabilities, which are back-propagated. 
Building on Excitation backprop,
\cite{Bargal_2018_CVPR} produce saliency maps for video RNNs.
 In our experiments, we produce spatial saliency maps using Grad-CAM, since it is efficient, easy to implement, widely used, and one of the saliency methods in \cite{adebayoneurips18} that passes the article's sanity checks.


Few works have been published with their focus on interpretability for video models (\cite{Feichtenhofer2018WhatRecognition}, \cite{Sigurdsson2017}, \cite{Huang_2018_CVPR} and \cite{GhodratiBMVC2018}). Other works have treated it, but with less extensive experimentation (\cite{Chattopadhyay2017Grad-CAM++:Networks}), while
mainly presenting a new spatiotemporal architecture
(\cite{Dwibedi_2018_CVPR_Workshops}, \cite{zhou2017temporalrelation}). We build on the work by
\cite{GhodratiBMVC2018}, where the aim is to measure a network's ability to model \textit{video time} directly, and not via the proxy task of action classification, which is most commonly seen. Three defining properties of video time are defined in the paper: temporal symmetry, temporal continuity and temporal causality, each accompanied by a measurable task.
The third property is measured using the classification accuracy on Something-something. 
An important contribution of ours is that we compare 3D CNNs and C-LSTMs, whereas \cite{GhodratiBMVC2018} compare 3D CNNs to  standard LSTMs. Their comparison can be argued as slightly unfair, as standard LSTM layers only take 1D input, and thus need to vectorize each frame, which removes some spatial dependencies in the pixel grid. \cite{BallasICLR16}, \cite{Dwibedi_2018_CVPR_Workshops} and  \cite{VideoLSTMLi2018} all use variants of convolutional RNNs, but only train them on CNN features. To the best of our knowledge, there has been no published convolutional RNNs trained on raw image data. This is crucial since information is lost when downsampling an image into CNN features, and we want to study networks that have sufficient degrees of freedom to learn temporal patterns from scratch.

Similar to our work, \cite{Dwibedi_2018_CVPR_Workshops} investigate the temporal modeling capabilities of convolutional gated recurrent units (ConvGRUs) trained on Something-something. The authors find that recurrent models perform well for the task, and present a qualitative analysis of the trained model's learned hidden states. For each class of the dataset, they obtain the hidden states of the network corresponding to the frames of one clip and display its nearest neighbors from other clips' per-frame hidden state representations. These hidden states had encoded information about the relevant frame ordering for the classes.
\cite{Sigurdsson2017} examine video architectures and datasets on a number of qualitative attributes. \cite{Huang_2018_CVPR} investigate how much the motion contributes to the classification performance of a video architecture.
To measure this, they vary the number of sub-sampled frames per clip to examine how much the accuracy changes as a result.

In a search-based precursor to our temporal mask experiments, \cite{Satkin-2010-10524} crop sequences temporally to obtain the most discriminative sub-sequence for a certain class.
The crop corresponding to the highest classification confidence is selected as the most discriminative sub-sequence. This selection is done using an exhaustive search for crops across all frames, which increases in complexity with the sequence length according to $\frac{\left|f\right|^2}{2}$, where $\left|f\right|$ is the number of frames. Our proposed method, however, is gradient-descent based and has a fixed number of iterations regardless of sequence length. Furthermore, our approach can identify more than one temporal sub-region in the sequence, in contrast to \cite{Satkin-2010-10524}.

\cite{Feichtenhofer2018WhatRecognition} present the first network centric interpretability work for video models. The authors investigate spatiotemporal features using activity maximization.
\cite{zhou2017temporalrelation} introduce the Temporal Relational Network (TRN), which learns temporal dependencies between frames through sampling the semantically relevant frames for a particular action class. The TRN module is put on top of a convolutional layer and consists of a fully connected network between the sampled frame features and the output.
Similar to \cite{Dwibedi_2018_CVPR_Workshops}, they perform temporal alignment of clips from the same class,
using
the frames considered most representative for the clip by the network. They verify the conclusion previously made by \cite{Xie2018ECCV}, that temporal order is crucial on Something-something and also investigate for which classes it is most important.

\section{Approach}
\label{section:approach}

\subsection{Temporal Masks}
The proposed temporal mask method aims to extend the interpretability of deep networks into the temporal dimension, utilizing meaningful perturbation of the input, as shown effective in the spatial dimension by \cite{MeaningFulPert}. When adopting this approach, it is necessary to define what constitutes a \textit{meaningful} perturbation. In the mentioned paper, a mask that blurs the input as little as possible is learned for a single image, while still maximizing the decrease in class score. Our proposed method applies this concept of a learned mask to the temporal dimension. The perturbation, in this setting, is a noise mask approximating either a 'freeze' operation, which removes motion data through time, or a 'reverse' operation that inverses the sequential order of the frames.
This way, we aim to identify which frames
are most critical for the network's classification decision.

The
temporal mask is defined as a vector of real numbers on the interval [0,1] with the same length as the input sequence. For the 'freeze' type mask, a value of 1 for a frame at index $t$ duplicates the value from the previous frame at $t-1$ onto the input sequence at $t$. The pseudocode for this procedure is given below.

\noindent
\begin{verbatim}
for i in maskIndicesExceptFirst do
  originalComponent := (1-mask[i])*originalInput[i]
  perturbedComponent := mask[i]*perturbedInput[i-1] 
  perturbedInput[i] := originalComponent + perturbedComponent
end for
\end{verbatim}




For the 'reverse' mask type, all indices of the mask \textbf{m} that are activated are first identified (threshold $0.1$). These indices are then looped through to find all contiguous sections, which are treated as sub-masks, $m_i$. For each sub-mask, the frames at the active indices in the sub-mask are reversed. For example (binary for clarity), an input indexed as $t_{1:16}$
perturbed with a mask with the value $[0,0,0,1,1,1,1,1,0,0,0,0,0,1,1,0]$ results in the sequence with frame indices $[1,2,3,8,7,6,5,4,9,10,11,12,13,15,14,16]$.

In order to learn the mask, we define a loss function (Eq. \ref{eq:maskLoss}) to be minimized using gradient descent, similar to the approach in \cite{MeaningFulPert}.

\begin{equation}
\label{eq:maskLoss}
\mathcal{L}=\lambda_1\lVert \mathbf{m} \rVert _1^1 + \lambda_2\lVert \mathbf{m} \rVert _{\beta}^{\beta} + F_c,
\end{equation}
\noindent where $\mathbf{m}$ is the mask expressed as a vector $m \in [0,1]^t$, $\lVert \cdot \rVert _{1}^{1}$ is the $L^1$ norm, $\lVert \cdot \rVert _{\beta}^{\beta}$ is the Total Variation (TV) norm, $\lambda_{1,2}$ are weighting factors, and $F_c$ is the class score given by the model for the perturbed input. The $L^1$ norm punishes long masks, in order to identify only the most important frames in the sequence.
The TV norm penalizes masks that are not contiguous.

This approach allows our method to automatically learn masks that identify one or several contiguous sequences in the input.
The mask is initialized centered in the middle of the sequence.
To keep the perturbed input class score differentiable with respect to the mask, the optimizer operates on a
real-valued
mask vector.
A sigmoid function is applied to the mask before using it for the perturbing operation in order to keep its values in the [0,1] range. 

 The ADAM optimizer is then used to learn the mask through 300 iterations of gradient descent. After the mask has converged, its sigmoidal representation is thresholded for visualisation purposes.

\subsection{Grad-CAM}

Grad-CAM (\cite{Selvaraju2017Grad-CAM:Localization}) is a method for producing visual explanations in the form of class-specific saliency maps for CNNs.
One saliency map, $L^{c}_t$, is produced for each image input based on the activations from $k$ filters, $A^k_{ij}$, at the final convolutional layer.
In order to adapt the method to sequences of images, the activations for all timesteps $t$ in the sequences are considered (Eq. \ref{eq:gradcam}).

\begin{equation}
\label{eq:gradcam}
    L^{c}_{ijt} = \sum_k w_{kt}^c A^k_{ijt} \quad; \qquad     w^c_{kt} = \frac{1}{Z}\sum_{ij}\frac{\partial F^c}{\partial A^k_{ijt}},
\end{equation}

\noindent where $Z$ is a normalizing constant and $F^c$ is the network output for the class $c$. Since the aim of the method is to identify which activations have the highest contribution to the class score, only positive values of the linear combination of activations are considered, as areas with negative values are likely to belong to other classes. 
By up-sampling these saliency maps to the resolution of the original input image, the aim is to examine what spatial data in specific frames contributed most to the predicted class. 
\section{Experiments}

\subsection{Datasets}
The Something-something dataset (\cite{Something-Something-v2}) contains over 220,000 sequences from 174 classes in a resolution of 224x224 pixels. The duration of the data is more than 200 hours, and the videos are recorded against varying backgrounds from different perspectives. The classes are action-oriented and object-agnostic. Each class is defined as performing some action with one or several arbitrary objects, such as \textit{closing something} or \textit{folding something}. This encourages the classifier to learn the action templates, since object recognition does not give enough information for the classifying task.
We train and validate according to the provided split. 

The KTH Actions dataset (\cite{KTHDATASET}) consists of 25 subjects performing six actions (\textit{boxing, waving, clapping, walking, jogging, running}) in four different settings, resulting in 2391 sequences, and a duration of almost three hours (160x120 pixels at 25 fps). They are filmed against a homogeneous background with the different settings exhibiting varying lighting, distance to the subject and clothing of the participants. For this dataset, we train on subjects 1-16 and evaluate on 17-25.

Both datasets have sequences varying from one to almost ten seconds. As 3D CNNs require input of fixed sequence length, all input sequences from both datasets are sub-sampled to cover the entire sequence in 16 frames (Something-something) and 32 frames (KTH Actions). The same set of sub-sampled frames is then used as input for both architectures.

\subsection{Architecture Details}
\label{sect:archi}
Both models were trained from scratch on each dataset, to ensure that the learned models were specific to the relevant task. Pre-training on Kinetics can increase performance, but for our experiments, the models should be trained on the temporal tasks presented by the Something-something dataset specifically. However, it can be noted that our I3D model reached comparable performance to another I3D trained from scratch on Something-something presented in the work of \cite{Xie2018ECCV}. Hyperparameters are listed in the supplementary material. Any remaining settings can be found in the code repository which can be found in the supplementary material. The code will be made public, in TensorFlow (\cite{tensorflow2015-whitepaper}) and PyTorch (\cite{PytorchNEURIPS2019_9015}).

\begin{figure}
\centering
\begin{subfigure}{.49\textwidth}
 \includegraphics[width=0.99\textwidth, align=c]{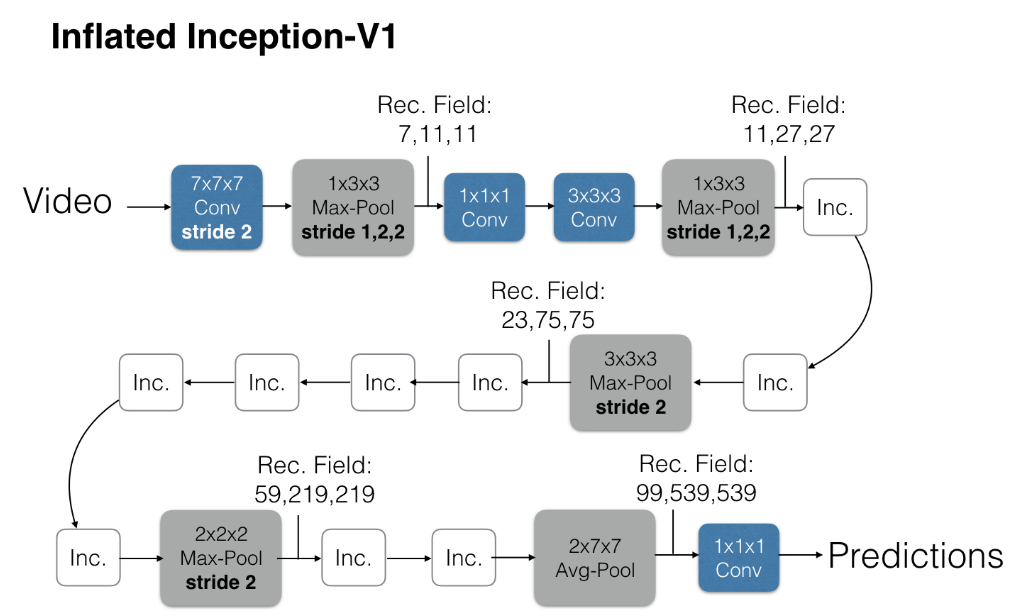}
\end{subfigure}
\begin{subfigure}{.49\textwidth}
 \includegraphics[width=0.99\textwidth, align=c]{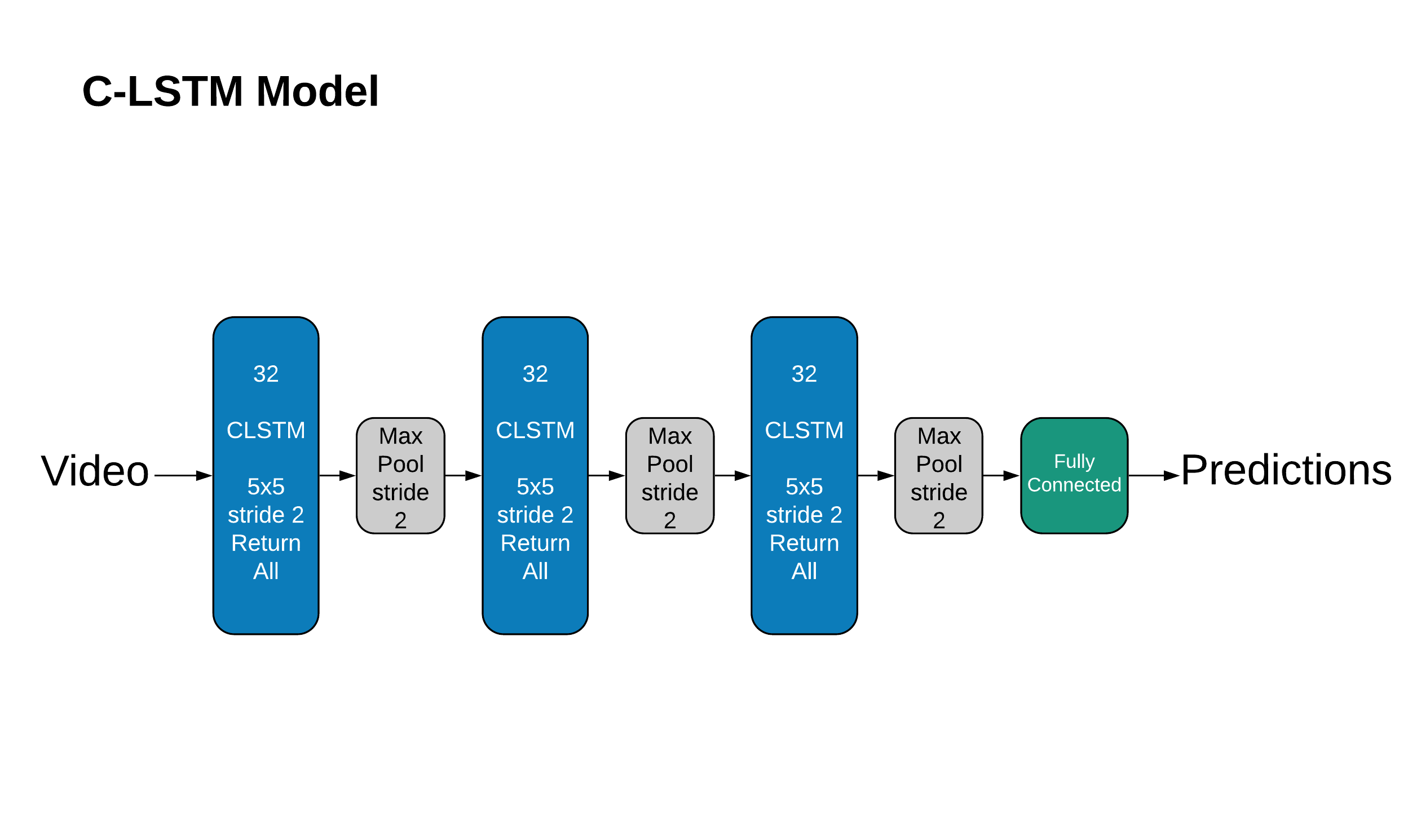}
\end{subfigure}
\caption{I3D network (figure from \cite{I3DRef}) and C-LSTM network (\textit{right}).}
\label{fig:ArchModels}
\end{figure}

I3D consists of three 3D convolutional layers, nine Inception modules and four max pooling layers (Fig. \ref{fig:ArchModels}). In the original setting, the temporal dimension of the input is down-sampled to $L/8$ frames by the final Inception module, where $L$ is the original sequence length. 
In order to achieve a higher temporal resolution in the produced Grad-CAM images, the strides of the first convolutional layer and the second max pooling layer are reduced to 1x2x2 in our code, producing $L/2$ activations in the temporal dimension. The Grad-CAM images are produced from the gradients of the class scores with respect to the final Inception module.

We have not found any published C-LSTMs trained on raw pixels, and thus conducted our own hyperparameter search for this model. The model was selected solely based on  classification performance; all feature investigations were conducted after this selection. The C-LSTM used for Something-something consists of three C-LSTM layers (two for KTH), each followed by batch normalization and max pooling layers. The convolutional kernels used for each layer had size 5x5 and stride 2x2 with 32 filters. The C-LSTM layers return the entire transformed sequence as input to the next layer.
When calculating the Grad-CAM maps for the C-LSTM, the final C-LSTM layer was used.

There is a substantial difference in the number of parameters for each model, with $12,465,614$ for I3D and $1,324,014$ and for the three-layer C-LSTM.
  Other variants of the C-LSTM with a larger number of parameters (up to five layers) were evaluated as well, but no significant increase in performance was observed. Also, due to the computational complexity of back-propagation through time (BPTT), the C-LSTM variants were significantly more time demanding to train than their I3D counterparts.

\subsection{Comparison Method}
To study the differences in the learned spatiotemporal features of the two models, we first compute the spatial saliency maps using Grad-CAM and the temporal masks using the proposed method. Once these are obtained for each model and dataset, we both examine them qualitatively and compute the quantitative metrics listed below. A 'blob' is defined as a contiguous patch within the Grad-CAM saliency map for one frame. The blobs were computed using the blob detection tool from OpenCV (\cite{opencv_library}). OS, FS and RS are the classification scores for the original input, and for the freeze and reverse perturbed input, respectively.

\begin{itemize}

\item \textbf{Blob count:} The average number of blobs (salient spatial areas, as produced by the Grad-CAM method), per frame.
    \item \textbf{Blob size:} The average size of one salient spatial area (blob), in pixels, computed across all detected blobs.
    \item \textbf{Center distance:} The average distance in pixels to the center of the frame for one blob, computed across all detected blobs.
    \item \textbf{Mask length:} The average number of salient frames per sequence, as produced by the temporal mask method.
    \item \textbf{Drop ratio:} The average ratio between the drop in classification score using the freeze and reverse perturbations, defined as $\frac{\text{OS}-\text{FS}}{\text{OS}-\text{RS}}$, across all sequences. 
    \item \textbf{Drop difference:} The average difference between the drop in classification score using the freeze and reverse perturbations, defined as $(\text{OS}-\text{FS})-(\text{OS}-\text{RS})$ (and equivalent to $\text{RS} - \text{FS}$), across all sequences.
\end{itemize}

We consider the difference and ratio between the freeze and reverse drops as the most relevant measures of how sensitive one model was for the
reverse perturbation.
FS and RS should not be compared in absolute numbers, since they depend on OS which might have been different for the two models. Moreover, using the same number of iterations for the optimization of the temporal mask, the two models typically reached different final losses (generally lower for I3D).

\section{Results}
For reference, the global validation F1-scores for both architectures and datasets are shown in Table \ref{table:f1res}. To emphasize the importance of temporal direction between the datasets, we first conduct a test where all the input validation sequences are entirely reversed.
On Something-something, both C-LSTM and I3D were affected drastically, while on KTH, both performed well.
Likely, this is because KTH's classes have distinct spatial features. As expected, Something-something is more time-critical than KTH. Overall, this shows that both models are indeed globally sensitive to temporal direction, when they need to be. In Sections \ref{section:qualsmth}-\ref{section:qualkth}, we examine in detail which spatiotemporal features are learned by the two models, and how they
differ from one another.

\begin{table}
\begin{tabular*}{\textwidth}{@{\extracolsep{\fill}}l|lll@{}}
\textbf{Model}             & \multicolumn{1}{c}{\textbf{KTH Actions} (Top-1)} & \multicolumn{1}{c}{\textbf{Smth-Smth} (Top-1)} & \multicolumn{1}{c}{\textbf{Smth-Smth} (Top-5)} \\ \hline
C-LSTM            & 0.84                           & 0.23                               & 0.48                                       \\
C-LSTM (rev.) & 0.78                           & 0.05                               & 0.17                                       \\ \hline
I3D               & 0.86                           & 0.43                               & 0.73                                       \\
I3D (rev.)    & 0.80                           & 0.09                               & 0.27                                       \\ \hline
\end{tabular*}

\caption{Validation F1-score per model on the two datasets. 'Rev.' indicates that the validation sequences were reversed at test time.}
\label{table:f1res}
\end{table}

\subsection{Interpretability Results on Something-something}
\label{section:qualsmth}
The less widely used C-LSTM architecture could not reach the same global performance as the state-of-the-art I3D (Table \ref{table:f1res}), which also has an order of magnitude more parameters.
The models were only compared on sequences from classes for which they had similar performance (Table \ref{table:F1Classes}).
 We include a variety of per-class F1-scores, ranging from approximately 0.1 to 0.9. All are, however, well above the random chance performance of $1/174 \approx 0.006$. The reason to include a variety of performance levels when studying the extracted features is to control for the general competence of the model. A well performing model might extract different features than a poor one.
 
In this section, we present an analysis of the Grad-CAM saliency maps and temporal masks generated for each architecture on the eleven classes. We evaluated the models on all validation sequences from these classes (1575 sequences in total). Quantitative results from the feature analysis are shown in Tables \ref{tab:gc}-\ref{tab:masks} and in Fig. \ref{fig:HistogramResults}. We display eight sample sequences in Figs. \ref{fig:smthsamples_1}-\ref{fig:smthsamples_2}, but include more qualitative examples in the supplementary material.

\begin{table}
\begin{tabular*}{\textwidth}{@{\extracolsep{\fill}}l|rr@{}}
\textbf{Class}             &  \multicolumn{1}{c}{\textbf{I3D}} & \multicolumn{1}{c}{\textbf{C-LSTM}}  \\ \hline
  \textit{burying something in something}           & 0.1                           & 0.06                                \\ \hline
  \textit{moving something and something away from each other}            & 0.76                           & 0.58                                \\ \hline
\textit{moving something and something closer to each other}          &  0.77                           & 0.57                                \\ \hline
\textit{moving something and something so they collide with each other}      &  0.16                           & 0.03                                \\ \hline
\textit{moving something and something so they pass each other}      &  0.37                           & 0.31                                \\ \hline
\textit{moving something up}         &  0.43                           & 0.40                                \\ \hline
\textit{pretending to take something from somewhere}         &  0.10                           & 0.07                                \\ \hline
\textit{turning the camera downwards while filming something}    &  0.67                           & 0.56                                \\ \hline
\textit{turning the camera left while filming something}    &  0.94                           & 0.79                               \\ \hline
\textit{turning the camera right while filming something}    & 0.91                           & 0.8                                \\ \hline
\textit{turning the camera upwards while filming something}        &  0.81                           & 0.73                                \\ \hline

\end{tabular*}
\caption{F1-score per class and model on the Something-something dataset.}
\label{table:F1Classes}
\end{table}

\subsubsection{Trends Regarding the Spatial Focus of the Two Models.}
We observe that the I3D generally focuses on contiguous, centered blobs, while the C-LSTM attempts to find relevant spatial features in multiple smaller areas (Table \ref{tab:gc}). Figs. \ref{fig:smthsamples_1}\subref{fig:away} and \ref{fig:smthsamples_1}\subref{fig:sotheypass} show examples of this, where I3D focuses on a single region covering both objects, while the C-LSTM has separate activations for the two objects, hands and the surface affected by the movement.

\begin{table}
\caption{Statistics for the Grad-CAM maps for each model on eleven classes from the validation set of Something-something (1575 sequences, 16 frames per sequence) and the whole test set of the KTH dataset (863 sequences, 32 frames per sequence). The 'blobs', i.e., the contiguous patches within each Grad-CAM map, were computed per frame, using the blob detection tool from OpenCV (\cite{opencv_library}).}
\begin{tabular*}{\textwidth}{@{\extracolsep{\fill}}l|lll@{}}
  \textbf{Model (Dataset)}     & \textbf{Blob count}     & \textbf{Blob size}       & \textbf{Center distance} \\ \hline
I3D (Smth-smth)    & $1.6 \pm 0.97$ & $33.7 \pm 19.6$ & $54.4 \pm 33.6$           \\
C-LSTM (Smth-smth) & $3.6 \pm 1.9$  & $26.7 \pm 24.5$ & $96.8 \pm 34.9$   \\ \hline
I3D (KTH)    & $ 1.1 \pm 0.5 $ & $ 44.0 \pm 18.7 $ & $ 44.6 \pm 19.4 $           \\ 
C-LSTM (KTH) & $ 32.6 \pm 15.1 $  & $ 5.8 \pm 7.0 $ & $ 49.9 \pm 22.4 $        \\  \hline
\end{tabular*}
\label{tab:gc}
\end{table}

We further find that the I3D has a bias of starting its focus around the middle of the frame (Figs. \ref{fig:smthsamples_1}-\ref{fig:smthsamples_2}), often even before the motion starts. 
This trend persists throughout the sequence, as the average distance to the center of the image for each blob in each frame is shorter for I3D (Table \ref{tab:gc}). The typical behavior for the C-LSTM is instead to remain agnostic until the action actually starts (e.g., Fig. \ref{fig:smthsamples_2}\subref{fig:downwards}).
In Fig. \ref{fig:smthsamples_2}\subref{fig:downwards}, the I3D maintains its foveal focus even after the green, round object is out of frame. In Fig. \ref{fig:smthsamples_2}\subref{fig:upwards}, the focus splits midway to cover both the moped and some features on the wall, while the C-LSTM focuses mainly on numerous features along the wall, as it usually does in classes where the camera turns. The C-LSTM also seems to pay more attention to hands appearing in the clips, rather than objects (Figs. \ref{fig:smthsamples_1}\subref{fig:away} and \ref{fig:smthsamples_1}\subref{fig:sotheypass}-\subref{fig:smthup2}). 

Fig. \ref{fig:HistogramResults} shows the normalized histograms of these spatial features.
The distributions for the two models differ significantly for all three measures.

\subsubsection{Trends of the Temporal Masks of the Two Models.}
\begin{table}
\caption{Statistics for the temporal masks of both models for both datasets (1575 sequences for Something-something and 863 sequences for KTH).}
\begin{tabular*}{\textwidth}{@{\extracolsep{\fill}}l|lll@{}}
 \textbf{Model (Dataset)}      & \textbf{Mask length}   & \textbf{Drop ratio} & \textbf{Drop diff.}  \\ \hline
I3D (Smth-smth)    & $6.2 \pm 3.3$ & $8.4 \pm 47$ & $0.2 \pm 0.3$                   \\
C-LSTM (Smth-smth) & $9.9 \pm 4.1$ & $2.6 \pm 6.9$ & $0.08 \pm 0.2$ \\ \hline 
I3D (KTH)    & $ 10.6 \pm 8.5 $ & $ 81.4 \pm 174$ & $ 0.57 \pm 0.34 $                   \\
C-LSTM (KTH) & $ 15.2 \pm 5.7 $ & $ 17.4 \pm 45.2 $ & $ 0.22 \pm 0.18 $           \\ \hline
\end{tabular*}
\label{tab:masks}
\end{table}

The quantitative results from the temporal mask experiments are shown in Table \ref{tab:masks}\footnote{For the drop ratio, if the denominator OS-RS $\le 0.001$, the sample was filtered out since its ratio would explode. The OS-FS $\le 0.001$ were also excluded for balance. When using $10^{-9}$ as threshold instead, the drop ratio results for Something-something were $215 \pm 6346$ (I3D) and $4.9 \pm 47.6$ (C-LSTM).}. We first note that the average temporal mask is shorter for the I3D. This suggests that it has learned to react to short, specific events in the sequences. As an example, its temporal mask in Fig. \ref{fig:smthsamples_1}\subref{fig:sotheypass} is active only on the frames where the objects first pass each other, and in Fig. \ref{fig:smthsamples_1}\subref{fig:closer}, it is active on the frames leading to the objects touching (further discussed in Section \ref{section:discussion}). Second, we note that the drop ratio and drop difference are generally higher for the I3D compared to C-LSTM (Table \ref{tab:masks}), suggesting that I3D is less sensitive to the reverse perturbation.

\begin{figure}
\begin{subfigure}{.99\textwidth}
 \includegraphics[width=0.99\textwidth, align=c]{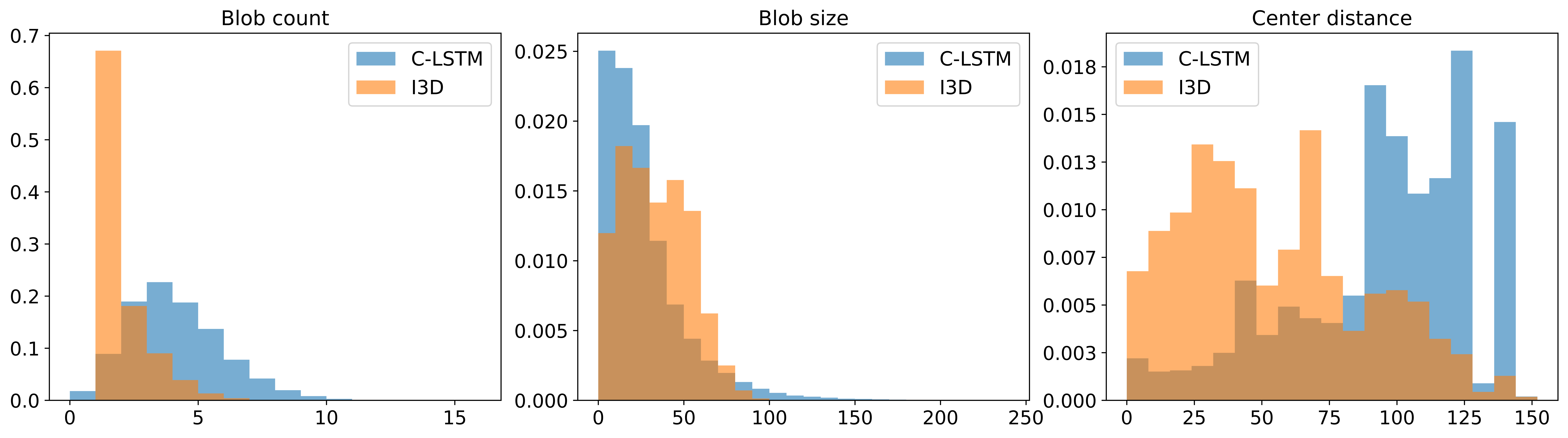}
\end{subfigure}
\begin{subfigure}{.99\textwidth}
 \includegraphics[width=0.99\textwidth, align=c]{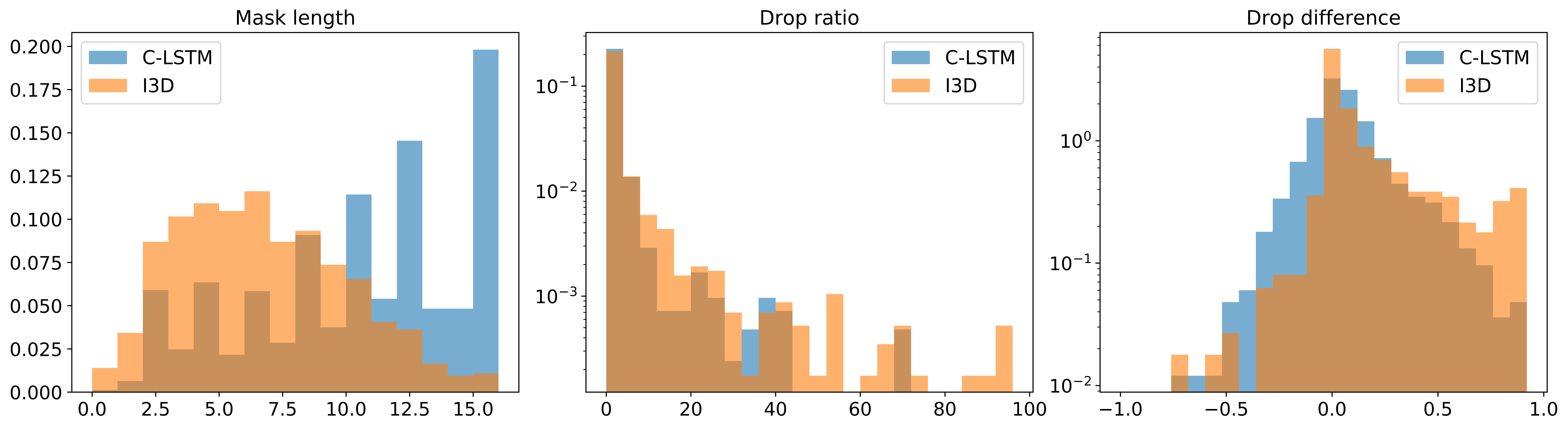}
\end{subfigure}
\caption{Normalized histogram results for the Grad-CAM and temporal mask analysis for the I3D (\textit{orange}) and C-LSTM (\textit{blue}) networks. The histograms correspond to the results in Tables \ref{tab:gc}-\ref{tab:masks}.}
\label{fig:HistogramResults}
\end{figure}

 The normalized histograms
 of the three measures are shown in 
 Fig. \ref{fig:HistogramResults}. The mask length distributions clearly have different means. For drop ratio and drop difference, the distributions have more overlap.
 A t-test conducted in Scipy (\cite{SciPy-NMeth2020}) of the difference in mean between the two models assuming unequal variance gives a p-value $ < 10^{-6}$ for both measures. We conclude that there is a significant difference in mean between the two models for drop ratio and drop difference.

\subsubsection{Class Ambiguity of the Something-something Dataset.}
The Something-something classes can be ambiguous (one class may contain another class) and, arguably, for some samples, incorrectly labeled. 
Examining the spatiotemporal features may give insight as to how the models handle these ambiguities. Fig. \ref{fig:smthsamples_1}\subref{fig:smthup2} shows a case of understandable confusion, where I3D answers \textit{taking one of many similar things on the table}. The surface seen in the image is a tiled floor, and the object is a transparent ruler. Once the temporal mask activates during the lifting motion in the last four frames, the Grad-CAM images show the model also focusing on two lines on the floor. These could be considered similar to the outline of the ruler, which could explain the incorrect classification. 
An example of ambiguous labeling can be seen for example in Fig. \ref{fig:smthsamples_1}\subref{fig:closer}, where I3D's classification is \textit{moving something and something so they collide with each other} and the C-LSTM predicts \textit{pushing something with something}. Although the two objects in the sequence do move closer to each other, they also touch at the
end, making both predictions technically correct.
\begin{figure}
\begin{subfigure}{.13\textwidth}
\begin{flushleft}
 OS: 0.994
 FS: 0.083 
 RS: 0.856 
 \end{flushleft}
 \end{subfigure}
    \begin{subfigure}{.72\textwidth}
        \animategraphics[width=125pt]{2}{Images/resultclips/36/43627/i3d/casefreeze43627_}{0}{15}
        \animategraphics[width=125pt]{2}{Images/resultclips/36/43627/clstm/casefreeze43627_}{0}{15}
        \caption{Moving something and something away from each other.}
        \label{fig:away}
    \end{subfigure}
\begin{subfigure}{.13\textwidth}
 \begin{flushright}
 OS: 0.312
 FS: 0.186 
 RS: 0.125 
 \end{flushright}
 \end{subfigure}

\begin{subfigure}{.13\textwidth}
\begin{flushleft}
OS: 0.547
FS: 0.028 
RS: 0.053 
CS: 0.186
P: 38
 \end{flushleft}
 \end{subfigure}
    \begin{subfigure}{.72\linewidth}
        \animategraphics[width=125pt]{2}{Images/resultclips/37/180193/i3d/casefreeze180193_}{0}{15}
        \animategraphics[width=125pt]{2}{Images/resultclips/37/180193/clstm/casefreeze180193_}{0}{15}
        \caption{Moving something and something closer to each other.}
        \label{fig:closer}
    \end{subfigure}
\begin{subfigure}{.13\textwidth}
 \begin{flushright}
 OS: 0.257
 FS: 0.079 
 RS: 0.122 
 CS: 0.002
 P: 135
 \end{flushright}
 \end{subfigure}
 
 \begin{subfigure}{.13\textwidth}
\begin{flushleft}
OS: 0.999
FS: 0.002 
RS: 0.414 
 \end{flushleft}
 \end{subfigure}
    \begin{subfigure}{.72\textwidth}
    \animategraphics[width=125pt]{2}{Images/resultclips/39/28170/i3d/casefreeze28170_}{0}{15}
    \animategraphics[width=125pt]{2}{Images/resultclips/39/28170/clstm/casefreeze28170_}{0}{15}
    \begin{center}\caption{Moving something and something so they pass each other.}\end{center}
    \label{fig:sotheypass}
    \end{subfigure}
\begin{subfigure}{.13\textwidth}
 \begin{flushright}
 OS: 0.788
 FS: 0.392 
 RS: 0.537 
 \end{flushright}
 \end{subfigure}
 
 \begin{subfigure}{.13\textwidth}
\begin{flushleft}
OS: 0.804
FS: 0.016 
RS: 0.667 
 \end{flushleft}
 \end{subfigure}
    \begin{subfigure}{.72\textwidth}
    \animategraphics[width=125pt]{2}{Images/resultclips/45/125857/i3d/casefreeze125857_}{0}{15}
    \animategraphics[width=125pt]{2}{Images/resultclips/45/125857/clstm/casefreeze125857_}{0}{15}
    \begin{center}\caption{Moving something up.}\end{center}
    \label{fig:smthup}
    \end{subfigure}
\begin{subfigure}{.13\textwidth}
 \begin{flushright}
 OS: 0.546
 FS: 0.121 
 RS: 0.764 
 \end{flushright}
 \end{subfigure}
 
 \begin{subfigure}{.13\textwidth}
\begin{flushleft}
OS: 0.685
FS: 0.003 
RS: 0.048 
CS: 0.001
P: 146
 \end{flushleft}
 \end{subfigure}
    \begin{subfigure}{.72\textwidth}
    \animategraphics[width=125pt]{2}{Images/resultclips/45/81536/i3d/casefreeze81536_}{0}{15}
    \animategraphics[width=125pt]{2}{Images/resultclips/45/81536/clstm/casefreeze81536_}{0}{15}
    \begin{center}\caption{Moving something up.}\end{center}
    \label{fig:smthup2}
    \end{subfigure}
\begin{subfigure}{.13\textwidth}
 \begin{flushright}
 OS: 0.221
 FS: 0.182 
 RS: 0.350 
 CS: 0.005
 P: 100
 \end{flushright}
 \end{subfigure}
 
 \begin{subfigure}{.13\textwidth}
\begin{flushleft}
OS: 0.284
FS: 0.003 
RS: 0.006 
 \end{flushleft}
 \end{subfigure}
    \begin{subfigure}{.72\textwidth}
    \animategraphics[width=125pt]{2}{Images/resultclips/81/46708/i3d/casefreeze46708_}{0}{15}
    \animategraphics[width=125pt]{2}{Images/resultclips/81/46708/clstm/casefreeze46708_}{0}{15}
    \begin{center}\caption{Pretending to take something from somewhere.}\end{center}
    \label{fig:pretendtake}
    \end{subfigure}
\begin{subfigure}{.13\textwidth}
 \begin{flushright}
 OS: 0.600
 FS: 0.167 
 RS: 0.088 
 CS: 0.004
 P: 27
 \end{flushright}
 \end{subfigure}
 \caption{\textbf{Best displayed in Adobe Reader where the figures can be played as videos, or in the supplementary material.} I3D (\textit{left}) and C-LSTM (\textit{right}) results for validation sequences from Something-something. The three columns show, from left to right, the original input, the Grad-CAM result, and the input as perturbed by the temporal freeze mask. The third column also visualizes when the mask is on (\textit{red}) or off (\textit{green}), with the current frame highlighted. \textit{OS}: original score (softmax output) for the guessed class, \textit{FS}: freeze score, \textit{RS}: reverse score, \textit{CS}: score for the ground truth class when there was a misclassification and \textit{P}: predicted label, if different from ground truth.}
\label{fig:smthsamples_1}
 \end{figure}
 
  \begin{figure}
 \begin{subfigure}{.13\textwidth}
\begin{flushleft}
OS: 1.000
FS: 0.001 
RS: 0.011 
 \end{flushleft}
 \end{subfigure}
    \begin{subfigure}{.72\textwidth}
\animategraphics[width=125pt]{2}{Images/resultclips/165/197549/i3d/casefreeze197549_}{0}{15}
    \animategraphics[width=125pt]{2}{Images/resultclips/165/197549/clstm/casefreeze197549_}{0}{15}
    \caption{Turning the camera downwards while filming something.}
    \label{fig:downwards}
    \end{subfigure}
\begin{subfigure}{.13\textwidth}
 \begin{flushright}
 OS: 0.158
 FS: 0.063 
 RS: 0.093 
 \end{flushright}
 \end{subfigure}
 
 \begin{subfigure}{.13\textwidth}
\begin{flushleft}
OS: 0.990
FS: 0.001 
RS: 0.000 
 \end{flushleft}
 \end{subfigure}
    \begin{subfigure}{.72\textwidth}
    \animategraphics[width=125pt]{2}{Images/resultclips/168/215115/i3d/casefreeze215115_}{0}{15}
    \animategraphics[width=125pt]{2}{Images/resultclips/168/215115/clstm/casefreeze215115_}{0}{15}
    \begin{center}\caption{Turning the camera upwards while filming something.}\end{center}
    \label{fig:upwards}
    \end{subfigure}
\begin{subfigure}{.13\textwidth}
 \begin{flushright}
 OS: 0.806
 FS: 0.177 
 RS: 0.181 
 \end{flushright}
 \end{subfigure}
 
    \caption{\textbf{Best displayed in Adobe Reader where the figures can be played as videos.}  Same structure as Fig. \ref{fig:smthsamples_1}.}
\label{fig:smthsamples_2}
\end{figure}


  \subsection{Interpretability Results on the KTH Actions Dataset}
  \label{section:qualkth}
For the KTH dataset, we make similar observations regarding temporal and spatial features. In Fig. \ref{fig:kthSamples}\subref{fig:handclapping}, we observe results for the class 'handclapping'. Interestingly, the mask of each model covers at least one entire cycle of the action. 
The reverse perturbation affects both models very little since one action cycle is symmetrical in time.
For the 'running' class (Fig. \ref{fig:kthSamples}\subref{fig:running}), we see that the temporal mask identifies the frames in which the subject is in-frame as the most salient for both models, with I3D placing more focus on the subject's legs.

\begin{figure}
    \begin{subfigure}{.13\textwidth}
     \begin{flushleft}
OS: 0.999
FS: 0.026
RS: 0.999
 \end{flushleft}
 \end{subfigure}
    \begin{subfigure}{.72\textwidth}
    \animategraphics[width=125pt]{2}{Images/resultclips/KTH/18_Handclapping_D3/i3d/casefreeze18_person18_handclapping_d3_1_}{0}{31}
    \animategraphics[width=125pt]{2}{Images/resultclips/KTH/18_Handclapping_D3/clstm/casefreezeperson18_handclapping_d3_}{0}{31}
    \begin{center}\caption{Handclapping, subject 18.}\end{center}
    \label{fig:handclapping}
    \end{subfigure}
\begin{subfigure}{.13\textwidth}
 \begin{flushright}
 OS: 0.996
 FS: 0.997
 RS: 0.996 
 \end{flushright}
 \end{subfigure}

\begin{subfigure}{.13\textwidth}
\begin{flushleft}
OS: 0.993\\
FS: 0.208\\
RS: 0.999 
 \end{flushleft}
 \end{subfigure}
    \begin{subfigure}{.72\linewidth}
    \animategraphics[width=125pt]{2}{Images/resultclips/KTH/25_Running_D3/i3d/casefreeze25_person25_running_d3_1_}{0}{31}
    \animategraphics[width=125pt]{2}{Images/resultclips/KTH/25_Running_D3/clstm/casefreezeperson25_running_d3_}{0}{31}
    \begin{center}\caption{Running, subject 25.}\end{center}
    \label{fig:running}
    \end{subfigure}
\begin{subfigure}{.13\textwidth}
 \begin{flushright}
 OS: 0.669
 FS: 0.339
 RS: 0.605 
 \end{flushright}
 \end{subfigure}

    \caption{\textbf{The figures can be displayed as videos in Adobe Reader.}
    Same structure as Fig. \ref{fig:smthsamples_1}.
    }
\label{fig:kthSamples}
\end{figure}

\subsection{Discussion}
\label{section:discussion}
As stated in Section \ref{section:intro}, 3D CNNs and C-LSTMs have fundamentally different ways of modeling time. In the following, we discuss two related observations: the shorter temporal masks of I3D and the fact that the classification scores after the freeze and reverse perturbations often are lower for I3D than for the C-LSTM.

For the I3D, all dimensions including the temporal axis of the input are progressively compressed through either convolutional strides or max pooling. The general understanding of CNNs are that later layers encode higher level features. In the deep video network examined in the work by \cite{Feichtenhofer2018WhatRecognition}, it is shown that the later layers have a preference for higher level actions. The representation that is input to the prediction layer in a 3D CNN has compressed high level motions or spatial relations through time to a shorter representation. The classification is then dependent on the presence or absence of these high level features in this representation. If perturbing the input would alter these critical high level features, the resulting prediction might be drastically affected.

For the C-LSTM, however,
hidden states resulting from the entire input sequence are
sent to the prediction layer. Ultimately, this means that it has a more temporally fine-grained feature space than its 3D CNN counterpart. We hypothesize that this is related to the two observed results.
Due to this fine-grained and enveloping temporal feature space,
the perturbation must remove larger sub-sequences from the data to obscure enough information through time to cause a large change in prediction score, possibly accounting for the longer temporal masks observed for C-LSTM. Furthermore, as we penalize the length of the mask 
during optimization, the resulting converged mask is often too short to fully bring down the classification score of the C-LSTM method. Examples of where the freeze
score is brought close to, or below, 0.1 are when the mask is nearly or fully active, as seen in Figs. \ref{fig:smthsamples_1}\subref{fig:closer}, \ref{fig:smthsamples_1}\subref{fig:smthup}  and \ref{fig:smthsamples_2}\subref{fig:downwards}.

\section{Conclusions and Future Work}
We have presented the first comparison of the spatiotemporal information used by 3D CNN and C-LSTM based models in action recognition.
We have presented indications that the difference in temporal modeling has consequences for what features the two models learn.
Using the proposed temporal mask method, we presented empirical evidence that I3D on average focuses on shorter and more specific sequences than the C-LSTM.
On average, our experiments showed that I3D also tends to focus on
fewer or a single contiguous spatial patch closer to the center of the image,
instead of smaller areas on several objects like the C-LSTM. Also, when comparing the effect of reversing the most salient frames or removing motion through 'freezing' them, the C-LSTM experiences a relatively larger decrease in prediction confidence than I3D upon reversal. We have also seen that the 
temporal mask is capable of identifying salient frames in sequences, such as one cycle of a repeated motion.

There is still much to explore in the patterns lying in temporal dependencies.
It would be of interest to extend the study to other datasets where temporal information is important, e.g., Charades (\cite{CharadesDataset}).
Other possible future work includes evaluating the effect of other noise types beyond 'freeze' and 'reverse'. We hope that this empirical study can guide future development and understanding of deep video models.

It is desirable that a model can be trained with as little data as possible. 3D CNNs do not represent video (time) in a physically sound way, treating it as a third spatial dimension. In our view, this is often made up for using large amounts of data and brute-force learning of its correlations, as most state-of-the-art video CNNs are from industry, trained on hundreds of GPUs, e.g., SlowFast (\cite{SlowFastFeichtenhofer_2019_ICCV}). For efficiency, it is important that the representation learned by the model should correspond to the world, and that variables that are uncorrelated in the world remain so in the model. With our evaluation framework it will be possible to gain further insight into what state-of-the-art video models have actually learned.


{\small
\bibliographystyle{iclr2020_conference}
\bibliography{iclr2020_conference}
}

\appendix
\section{Appendix}
\subsection{Training Hyperparameters}
Tables \ref{table:trainingparams}-\ref{table:maskparams} present the hyperparameters used for the training of the two models on the two datasets, and for the learning of the temporal masks.

\begin{table}[h!]
\caption{Hyperparameters used for training the two models on each dataset.}
\begin{tabular}{lrrrrr}
                  & \multicolumn{5}{c}{Hyperparameter}                                                                                                                             \\ \cline{2-6} 
Model (Dataset)   & \multicolumn{1}{c}{Dropout Rate} & \multicolumn{1}{c}{Weight Decay} & \multicolumn{1}{c}{Optimizer} & \multicolumn{1}{l}{Epochs} & \multicolumn{1}{l}{Momentum} \\ \hline
I3D (Smth-smth)   & 0.5                              & 0                                & ADAM                          & 13                         & -                            \\ \hline
\rowcolor[HTML]{EFEFEF} 
I3D (KTH)         & 0.7                              & 5E-5                             & ADAM                          & 30                         & -                            \\ \hline
C-LSTM (Smth-smth) & 0.0                              & 0                                & SGD                           & 105                        & 0.2                          \\ \hline
\rowcolor[HTML]{EFEFEF} 
C-LSTM (KTH)       & 0.5                              & 1E-4                             & SGD                           & 21                         & 0.2                          \\ \hline
\end{tabular}
\label{table:trainingparams}
\end{table}

\begin{table}[h!]
\caption{Hyperparameters used for optimizing the temporal mask.}
\begin{tabular}{lrrrrrr}
                  & \multicolumn{6}{c}{Hyperparameter}                                                                                                                             \\ \cline{2-7} 
Dataset   & \multicolumn{1}{c}{$\lambda_1$} & \multicolumn{1}{c}{$\lambda_2$} &
\multicolumn{1}{c}{$\beta$} & \multicolumn{1}{c}{Optimizer} & \multicolumn{1}{l}{Iterations} & \multicolumn{1}{l}{Learning rate} \\ \hline
Smth-smth   & 0.01                              & 0.02 & 3                               & ADAM                          & 300                         & 0.001                            \\ \hline
\rowcolor[HTML]{EFEFEF} KTH         & 0.02                              & 0.04   & 3                          & ADAM                          & 300                         & 0.001                            \\ \hline
\end{tabular}
\label{table:maskparams}
\end{table}

\subsection{Further Sequence Examples: Something-something}

On the project website\footnote{\url{https://interpreting-video-features.github.io/}}, we display results for 22 additional randomly selected sequences (two from each class) from the Something-something dataset. As mentioned in the main article, we selected eleven classes for our experiments where the two models had comparable performance. The four classes not appearing in the main article are (I3D F1-score/C-LSTM F1 score): \textit{moving something and something so they collide with each other} (0.16/0.03), \textit{burying something in something} (0.1/0.06), \textit{turning the camera left while filming something} (0.94/0.79) and \textit{turning the camera right while filming something} (0.91/0.8).

\section{Something-something classes}

 "Approaching something with your camera":"0",
  "Attaching something to something":"1",
  "Bending something so that it deforms":"2",
  "Bending something until it breaks":"3",
  "Burying something in something":"4",
  "Closing something":"5",
 "Covering something with something":"6",
 "Digging something out of something":"7",
 "Dropping something behind something":"8",
 "Dropping something in front of something":"9",
 "Dropping something into something":"10",
 "Dropping something next to something":"11",
 "Dropping something onto something":"12",
 "Failing to put something into something because something does not fit":"13",
 "Folding something":"14",
 "Hitting something with something":"15",
 "Holding something":"16",
 "Holding something behind something":"17",
 "Holding something in front of something":"18",
 "Holding something next to something":"19",
 "Holding something over something":"20",
 "Laying something on the table on its side, not upright":"21",
 "Letting something roll along a flat surface":"22",
 "Letting something roll down a slanted surface":"23",
 "Letting something roll up a slanted surface, so it rolls back down":"24",
 "Lifting a surface with something on it but not enough for it to slide down":"25",
 "Lifting a surface with something on it until it starts sliding down":"26",
 "Lifting something up completely without letting it drop down":"27",
 "Lifting something up completely, then letting it drop down":"28",
 "Lifting something with something on it":"29",
 "Lifting up one end of something without letting it drop down":"30",
 "Lifting up one end of something, then letting it drop down":"31",
 "Moving away from something with your camera":"32",
 "Moving part of something":"33",
 "Moving something across a surface until it falls down":"34",
 "Moving something across a surface without it falling down":"35",
 "Moving something and something away from each other":"36",
 "Moving something and something closer to each other":"37",
 "Moving something and something so they collide with each other":"38",
 "Moving something and something so they pass each other":"39",
 "Moving something away from something":"40",
 "Moving something away from the camera":"41",
 "Moving something closer to something":"42",
 "Moving something down":"43",
 "Moving something towards the camera":"44",
 "Moving something up":"45",
 "Opening something":"46",
 "Picking something up":"47",
 "Piling something up":"48",
 "Plugging something into something":"49",
 "Plugging something into something but pulling it right out as you remove your hand":"50",
 "Poking a hole into some substance":"51",
 "Poking a hole into something soft":"52",
 "Poking a stack of something so the stack collapses":"53",
 "Poking a stack of something without the stack collapsing":"54",
 "Poking something so it slightly moves":"55",
 "Poking something so lightly that it doesn't or almost doesn't move":"56",
 "Poking something so that it falls over":"57",
 "Poking something so that it spins around":"58",
 "Pouring something into something":"59",
 "Pouring something into something until it overflows":"60",
 "Pouring something onto something":"61",
 "Pouring something out of something":"62",
 "Pretending or failing to wipe something off of something":"63",
 "Pretending or trying and failing to twist something":"64",
 "Pretending to be tearing something that is not tearable":"65",
 "Pretending to close something without actually closing it":"66",
 "Pretending to open something without actually opening it":"67",
 "Pretending to pick something up":"68",
 "Pretending to poke something":"69",
 "Pretending to pour something out of something, but something is empty":"70",
 "Pretending to put something behind something":"71",
 "Pretending to put something into something":"72",
 "Pretending to put something next to something":"73",
 "Pretending to put something on a surface":"74",
 "Pretending to put something onto something":"75",
 "Pretending to put something underneath something":"76",
 "Pretending to scoop something up with something":"77",
 "Pretending to spread air onto something":"78",
 "Pretending to sprinkle air onto something":"79",
 "Pretending to squeeze something":"80",
 "Pretending to take something from somewhere":"81",
 "Pretending to take something out of something":"82",
 "Pretending to throw something":"83",
 "Pretending to turn something upside down":"84",
 "Pulling something from behind of something":"85",
 "Pulling something from left to right":"86",
 "Pulling something from right to left":"87",
 "Pulling something onto something":"88",
 "Pulling something out of something":"89",
 "Pulling two ends of something but nothing happens":"90",
 "Pulling two ends of something so that it gets stretched":"91",
 "Pulling two ends of something so that it separates into two pieces":"92",
 "Pushing something from left to right":"93",
 "Pushing something from right to left":"94",
 "Pushing something off of something":"95",
 "Pushing something onto something":"96",
 "Pushing something so it spins":"97",
 "Pushing something so that it almost falls off but doesn't":"98",
 "Pushing something so that it falls off the table":"99",
 "Pushing something so that it slightly moves":"100",
 "Pushing something with something":"101",
 "Putting number of something onto something":"102",
 "Putting something and something on the table":"103",
 "Putting something behind something":"104",
 "Putting something in front of something":"105",
 "Putting something into something":"106",
 "Putting something next to something":"107",
 "Putting something on a flat surface without letting it roll":"108",
 "Putting something on a surface":"109",
 "Putting something on the edge of something so it is not supported and falls down":"110",
 "Putting something onto a slanted surface but it doesn't glide down":"111",
 "Putting something onto something":"112",
 "Putting something onto something else that cannot support it so it falls down":"113",
 "Putting something similar to other things that are already on the table":"114",
 "Putting something that can't roll onto a slanted surface, so it slides down":"115",
 "Putting something that can't roll onto a slanted surface, so it stays where it is":"116",
 "Putting something that cannot actually stand upright upright on the table, so it falls on its side":"117",
 "Putting something underneath something":"118",
 "Putting something upright on the table":"119",
 "Putting something, something and something on the table":"120",
 "Removing something, revealing something behind":"121",
 "Rolling something on a flat surface":"122",
 "Scooping something up with something":"123",
 "Showing a photo of something to the camera":"124",
 "Showing something behind something":"125",
 "Showing something next to something":"126",
 "Showing something on top of something":"127",
 "Showing something to the camera":"128",
 "Showing that something is empty":"129",
 "Showing that something is inside something":"130",
 "Something being deflected from something":"131",
 "Something colliding with something and both are being deflected":"132",
 "Something colliding with something and both come to a halt":"133",
 "Something falling like a feather or paper":"134",
 "Something falling like a rock":"135",
 "Spilling something behind something":"136",
 "Spilling something next to something":"137",
 "Spilling something onto something":"138",
 "Spinning something so it continues spinning":"139",
 "Spinning something that quickly stops spinning":"140",
 "Spreading something onto something":"141",
 "Sprinkling something onto something":"142",
 "Squeezing something":"143",
 "Stacking number of something":"144",
 "Stuffing something into something":"145",
 "Taking one of many similar things on the table":"146",
 "Taking something from somewhere":"147",
 "Taking something out of something":"148",
 "Tearing something into two pieces":"149",
 "Tearing something just a little bit":"150",
 "Throwing something":"151",
 "Throwing something against something":"152",
  "Throwing something in the air and catching it":"153",
  "Throwing something in the air and letting it fall":"154",
  "Throwing something onto a surface":"155",
  "Tilting something with something on it slightly so it doesn't fall down":"156",
  "Tilting something with something on it until it falls off":"157",
  "Tipping something over":"158",
  "Tipping something with something in it over, so something in it falls out":"159",
  "Touching (without moving) part of something":"160",
  "Trying but failing to attach something to something because it doesn't stick":"161",
  "Trying to bend something unbendable so nothing happens":"162",
  "Trying to pour something into something, but missing so it spills next to it":"163",
  "Turning something upside down":"164",
  "Turning the camera downwards while filming something":"165",
  "Turning the camera left while filming something":"166",
  "Turning the camera right while filming something":"167",
  "Turning the camera upwards while filming something":"168",
  "Twisting (wringing) something wet until water comes out":"169",
  "Twisting something":"170",
  "Uncovering something":"171",
  "Unfolding something":"172",
  "Wiping something off of something":"173"
 
 \section{KTH actions classes}
 
 "Boxing": "0", "Handclapping": "1", "Handwaving":"2", "Jogging":"3", "Running":"4", "Walking":"5"

\end{document}